\documentclass[journal, twocolumn]{IEEEtran}
\usepackage{url,algorithm,algpseudocode,amsfonts, amsmath, graphicx, stmaryrd, subfigure, cite, color, booktabs, multirow}

\allowdisplaybreaks

\newcommand*{\myfont}{\fontfamily{qcr}\footnotesize\selectfont}
\DeclareTextFontCommand{\textmyfont}{\myfont}

\def\bth{\boldsymbol{\theta}}

\def\({\left(}
\def\){\right)}
\def\[{\left[\,}
\def\]{\,\right]}

\def\0{\boldsymbol{0}}
\def\1{\boldsymbol{1}}
\def\a{\mathbf{a}}
\def\A{\mathbf{A}}

\def\b{\mathbf{b}}
\def\B{\mathbf{B}}

\def\c{\mathbf{c}}
\def\C{\mathbf{C}}

\def\g{\mathbf{g}}
\def\G{\mathbf{G}}

\def\n{\mathbf{n}}
\def\N{\mathbf{N}}

\def\bR{\mathbb{R}}

\def\u{\mathbf{u}}

\def\x{\mathbf{x}}
\def\X{\mathbf{X}}

\def\Y{\mathbf{Y}}
\def\z{\mathbf{z}}
\def\Z{\mathbf{Z}}

\newtheorem{remark}{Remark}

\def\tG{\underline{\G}}
\def\tX{\underline{\X}}

\def\tZ{\underline{\Z}}
\def\cP{\mathcal{P}}
\def\cD{\mathcal{D}}
\def\suchthat{\mathrm{s.\,t.}}

\begin{document}

%\subtitle{Subject Section}

\title{REP: Predicting the Time-Course of Drug Sensitivity}
\author{Cheng Qian, Amin Emad, and Nicholas D. Sidiropoulos}

%This work was supported in part by the National Science Foundation under project NSF IIS-1704074

\maketitle

\begin{abstract}
	\textbf{Motivation:} 
	The biological processes involved in a drug's mechanisms of action are oftentimes dynamic, complex and difficult to discern. Time-course gene expression data is a rich source of information that can be used to unravel these complex processes, identify biomarkers of drug sensitivity and predict the response to a drug. However, the majority of previous work has not fully utilized this temporal dimension. In these studies, the gene expression data is either considered at one time-point (before the administration of the drug) or two time-points (before and after the administration of the drug). This is clearly inadequate in modeling dynamic gene-drug interactions, especially for applications such as long-term drug therapy.
	
	\textbf{Results:} In this work, we present a novel REcursive Prediction (REP) framework for drug response prediction by taking advantage of time-course gene expression data. 
	Our goal is to predict drug response values at every stage of a long-term treatment, given the expression levels of genes collected in the previous time-points. 
	To this end, REP employs a built-in recursive structure that exploits the intrinsic time-course nature of the data and integrates past values of drug responses for subsequent predictions. It also incorporates tensor completion that can not only alleviate the impact of noise and missing data, but also predict unseen gene expression levels (GELs). These advantages enable REP to estimate drug response at any stage of a given treatment from some GELs measured in the beginning of the treatment. Extensive experiments on a dataset corresponding to 53 multiple sclerosis patients treated with interferon are included to showcase the effectiveness of REP.
\end{abstract}

\section{Introduction}
Prediction of drug response based on patients' clinical and molecular features is a major challenge in personalized medicine. Great effort has been devoted to identify molecular biomarkers of drug sensitivity and to develop computational models to predict drug response based on these biomarkers. 
Gene expression data is one of the most commonly used data type in these studies, due to their high predictive ability, and numerous methods have been proposed for drug response prediction based on gene expression data \cite{costello2014community,suphavilai2018predicting,qian2019gene,zhang2018novel,mcdermott2018cancer,bar2012studying,fukushima2019elastic}. 	
However, many existing methods only use basal gene expression data (i.e., gene expression values before administration of the drug) and hence can only capture the influence of the steady state of the cells on their response to a drug. For example, the authors of \cite{costello2014community} analyzed 44 drug response prediction methods that employed gene expression profiles of breast cancer cell lines taken before treatment to predict dose-response values, e.g., $\mathrm{GI}_{50}$--the concentration that inhibited cell growth by 50\% after 72 hours of treatment. Similar work can also be found in \cite{suphavilai2018predicting,qian2019gene,zhang2018novel}, which only incorporate gene expression data from a single time-point.
A collection of temporal gene expression profiles of samples over a series of time-points during the course of a biological process can provide more insights than a single (or two) time-point(s) \cite{luan2003clustering}. Therefore, developing algorithms that can predict the drug response over time using time-course gene data is of great interest. 

With the advancement of gene sequencing technologies, collecting gene expression data over multiple time-points and their matched drug response values is now feasible. In parallel with these technological developments, there has been growing interest in the application of machine learning methods to analyze the time-course gene expression data. In \cite{baranzini2004transcription}, the authors proposed an integrated Bayesian inference system to select genes for drug response classification from time-course gene expression data. However, the method only uses the data from the first time-point, and hence does not benefit from the additional temporal information. Lin et al., \cite{lin2008alignment} presented a Hidden Markov model (HMM)- based classifier, in which the HMM had fewer states than time points to align different patient response rates. This discriminative HMM classifier enabled distinguishing between good/bad responders. Nevertheless, choosing the number of states for this HMM is a major practical issue. In addition, this method cannot handle missing data and it requires the full knowledge of GELs in all time-points {\em a prior}i. This implies that the HMM may not be able to predict drug response at multiple stages in future time points, since the corresponding GELs are not measurable.

The time-course gene expression data contains the GELs of different patients over a series of time points, which can be indexed as patient-gene-time and represented as a three-dimensional tensor. Motivated by this, several tensor decomposition  based algorithms have been proposed. For example, Taguchi \cite{taguchi2017identification} employed tensor decomposition to identify drug target genes using time-course gene expression profiles of human cell lines. Li and Ngom \cite{li2010non} proposed a higher-order nonnegative matrix factorization (HONMF) tool for classifying good or bad responders from a latent subspace corresponding to patients learned from HONMF. One limitation of this work is that the latent subspace may not have discriminative ability in classifying patients, since it is learned without accounting for the class-label information. Moreover, this method simply discards samples with missing values, causing unnecessary information loss.

Recently, Fukushima \emph{et al.,} \cite{fukushima2019elastic} developed an algorithm for joint gene selection and drug response prediction for time-course data. The method uses Elastic-Net (EN) to select a set of genes that show discrimination of patients' drug responses throughout the treatment. The selected genes are then passed to a logistic regression (LR) classifier for drug response prediction. But in real applications, due to the existence of noise and missing values in the data, finding discriminative genes for all patients may be difficult. In fact, several studies have shown that it is more viable to find genes that have consistent discrimination in a subset of samples along the time series  \cite{zhao2005tricluster,jiang2004mining,mikalsen2018time}. 
Therefore, relying only on discriminative gene selection but without modifying classification algorithms may not achieve satisfactory performance.

In this paper, we take a different approach for time-course drug response prediction. We hypothesize that a patient's drug response at a given time-point can be inferred from the response at a previous time point. This means that not only the GELs but also the past response results can be integrated to identify the drug response for a subsequent time point. We develop a REcursive Prediction (REP) algorithm to predict the drug response of samples using their time-course gene expression data and their drug response at previous time-points. REP has a built-in recursive structure that exploits the intrinsic time-course nature of the data through integrating past drug responses for subsequent prediction. In other words, in REP, not only the GELs but also the past drug responses are treated as features for drug response prediction. Furthermore, by taking into consideration the intrinsic tensor structure of the time-course gene expression data and leveraging identifiability of low-rank tensors, REP can alleviate the noise corruption in GEL measurements, complete missing GELs and even predict GELs for subsequent time points. These features enable REP to evaluate drug response at any stage of a given treatment from some GELs measured in the beginning of a treatment. Experiments on real data are included to demonstrate the effectiveness of the REP algorithm.

\section{Method}
\begin{figure*}[t]
    \centering
	\subfigure[Pre-processing]{\label{fig:Pre-processing} \includegraphics[width=0.85\linewidth]{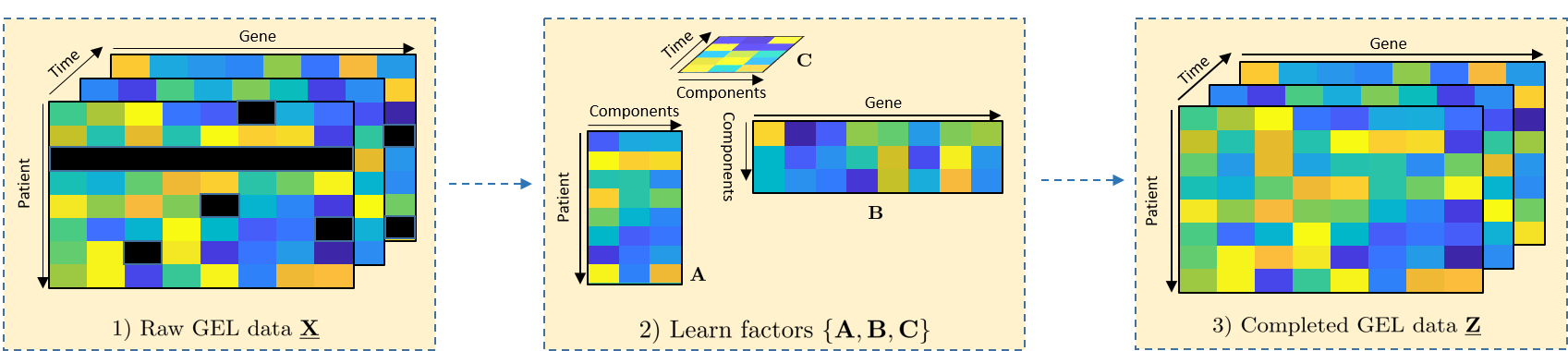}} 
	\subfigure[Training]{\label{fig:training} \includegraphics[width=0.3\linewidth]{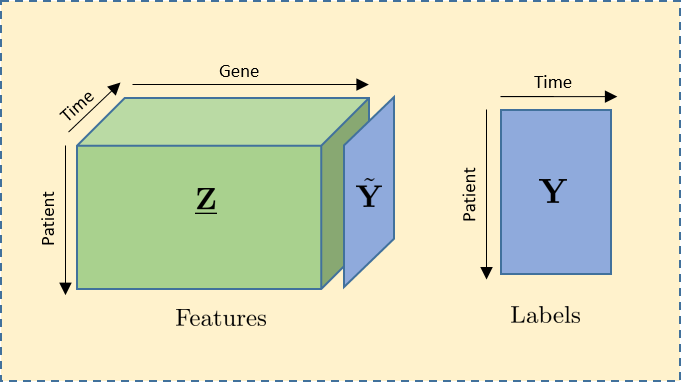}}
	\subfigure[Prediction]{\label{fig:testing} \includegraphics[width=0.54\linewidth,clip=true]{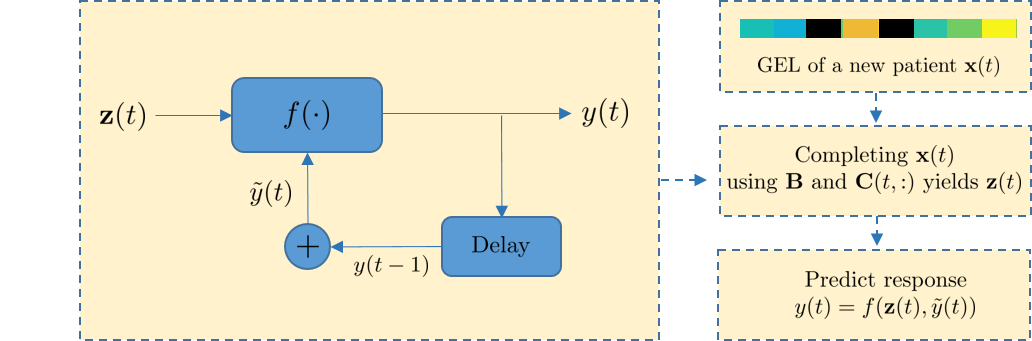}}
	\caption{Sketch view of the proposed method. In (a), Step 1) shows the raw data $\tX$ with missing values marked as `black'; Step 2) shows the low-rank tensor factorization; Step 3) is the missing completion, where $\tZ = \cP_{\Omega}(\tX) + \cP_{\Omega^c}(\tG)$. In (b), it shows the composition of training data: features and labels. In (c), it shows the prediction for new patient at a specific time $t$.}\label{fig:sketchview}
\end{figure*}

Fig. \ref{fig:sketchview} sketches the idea behind the proposed REP algorithm, where the subfigures \ref{fig:Pre-processing}--\ref{fig:testing} show the pre-processing, model training and prediction of our method, respectively. The tensor structure of time-course gene expression data is shown in Fig. \ref{fig:Pre-processing}.
In the following, we explain them in more detail.

\subsection{Pre-processing}\label{ntc}

One major issue in using gene expression data for drug response prediction is the existence of missing values. To overcome this problem, we first impute the missing values during pre-processing. Various methods have bene previously suggested for handling missing values, such as median-imputation \cite{moorthy2014review} and nearest neighbor imputation \cite{dudoit2002comparison,li2010missing}. 
We employ a low-rank tensor model to fit the time-course gene expression dataset such that the missing values can be completed. Our supporting hypothesis is that genes never function in an isolated way, but oftentimes groups of genes interact together to maintain the complex biological process, which results in correlation in the GEL data \cite{kapur2016gene}. We note that our low-rank tensor model suggests three factors that uniquely determine the values of GELs, i.e., the factors corresponding to patient, gene and time, respectively (see Fig. \ref{fig:sketchview}). 
As we will see later, our model allows us to estimate the variation of GEL over time from a set of initial GEL measurements; these estimated values are then used to predict the time-course of drug response. 

Towards this goal, we first assume\footnote{For high-enough but finite $F$, any patient-gene-time dataset can be expressed this way. See \cite{TSPOP} for a tutorial overview of tensor rank decomposition.} that each GEL is represented as a summation of $F$ triple products from the latent factors of patient, gene and time, respectively.
Let us denote $g_{ijk}$ as the $j$th GEL of patient $i$ recorded at time $k$. Based on our assumption, we have
\begin{align}\label{xijk}
g_{ijk} =\sum_{f=1}^{F} a_{if}b_{jf}c_{kf}
\end{align}
where $a_{if}$, $b_{jf}$ and $c_{kf}$ are the latent factors of patient, gene and time, respectively. 
Suppose that there are $J$ genes measured over $K$ time points. By varying the indices $j$ and $k$ in \eqref{xijk}, the expression of the genes in all the time-points in patient $i$ can be represented as
\begin{align}\label{Xj}
\G_{i} = \B\cD_i(\A)\C^T \in \bR^{J\times K}
\end{align}
where $\A\in\bR^{I\times F}$, $\B\in\bR^{J\times F}$, $\C\in\bR^{K\times F}$. In this equation, $\cD_i(\A)$ represents a diagonal matrix holding the $i$th row of $\A$ as the main diagonal, which is a latent representation of the $i$th patient. 
We use $a_{if}$ to represent the $(i; f)$-entry of $\A$, $b_{jf}$ to represent the $(j; f)$-entry of $\B$ and $c_{kf}$ to represent the $(k; f)$-entry of $\C$.

Assume that there are $I$ patients in the training set. After collecting $\{\G_1,\cdots,\G_I\}$, we stack them in parallel along the patient-axis, which results in a GEL tensor that takes the form of
\begin{align}
\tG := \left\llbracket \A,\B,\C \right\rrbracket  = \sum_{f=1}^F\a_f\circ\b_f\circ\c_f \in\bR^{I\times J\times K}
\end{align}
where $\circ$ is the outer product and $\a_f$ is the $f$th column of $\A$, and likewise for $\b_f$ and $\c_f$. 
Here, we assume that $\tG$ is the noiseless GEL data and $\tX$ is the corresponding noisy data with missing values. The relationship between $\tG$ and $\tX$ is described as
\begin{align}
\cP_\Omega(\tX) = \cP_\Omega(\tG) + \cP_\Omega(\underline{\N})
\end{align}
where $\underline{\N}$ is the noise in the data, $\Omega$ is the index set of the observed GELs in $\tX$, and $\cP_{\Omega}$ is the operator that keeps the entries in $\Omega$ and zeros out the others.

The model in \eqref{Xj} indicates that the gene and time factors (i.e., $\B$ and $\C$) are identical for different patients, and the variability among patients is captured by $\A$. In other words, given $\B$ and $\C$, each row of the patient factor matrix $\A$ uniquely determines the GELs of the corresponding patient. 
As we will see later, our model is able to predict unseen GELs, which also enables to prescreen the drug response for different stages of a treatment. 

Assuming nonnegative GELs\footnote{Due to some preprocessing steps such as z-score normalization, the GEL values can be negative. To facilitate our method, we undo these preprocessing steps or use the raw dataset. }, we can use nonnegative tensor factorization to compute missing GEL values:
\begin{align}\label{problem}
\min_{\A,\B,\C,\tG} &~ \left\|\tG - \left\llbracket \A,\B,\C \right\rrbracket\right\|_F^2 + \lambda\(\|\A\|_F^2 + \|\B\|_F^2 + \|\C\|_F^2\)\notag\\
\suchthat\; &\;\cP_{\Omega}(\tG) = \cP_{\Omega}(\tX), \A\geq 0, \B\geq 0, \C\geq 0
\end{align} 
where many sophisticated algorithms are applicable to optimize \eqref{problem}, e.g., block coordinate descent \cite{xu2013block,TSPOP}. Intuitively, \eqref{problem} seeks to identify the lowest rank solution $\G$ that best matches the observations $\tX$. The regularization terms are added to further encourage low rank and prevent over-fitting.
When \eqref{problem} is solved, we complete the GEL data through
\begin{align}
\tZ = \cP_{\Omega}(\tX) + \cP_{\Omega^c}(\tG)
\end{align}
where $\Omega^c$ contains the indices of missing values in $\tX$.

\subsection{Training}
The effects of drugs are usually cumulative over time \cite{craig1998pharmacokinetic}, i.e., drug doses taken in the past will affect the current response. This implies that the drug response in the past time-points may help predict the current response. Based on this hypothesis, we propose a recursive prediction algorithm, henceforth referred to as REP for simplicity, which enables to integrate past drug response records with gene expression values for subsequent drug response predictions. Fig. \ref{fig:testing} shows an overview of REP's pipeline, where drug responses $\{y_0,\cdots,y(t-1)\}$ in the previous time stages are integrated with the gene expression information $\z_{t}$ for predicting the current response $y(t)$. Here, we accumulate the historical response as
\begin{align}\label{yt}
\tilde{y}(t) = \sum_{i=0}^{t-1}y(i)
\end{align}
which is then fed back as an input feature for subsequent drug response prediction.
The intuition behind \eqref{yt} is that if the responses at adjacent time stages are consistent, then the response at the next stage is more likely to continue the same pattern instead of going in the opposite direction. If the responses at two adjacent time stages are opposite, i.e., one is good and the other one is poor, their effect will be canceled out in \eqref{yt}. In this case, the predictor will focus more on the GELs. 

Mathematically, we have
\begin{equation}
y_{i,t} = f(\z_{i,t},\tilde{y}_{i,t})
\end{equation}
where $\tilde{y}_{i,t}$ denotes the accumulated historical responses of patient $i$ at time $t$, $f(\cdot)$ is the predictor which can be trained by minimizing the following cost function
\begin{align}
L(\bth) =  \frac{1}{IK}\sum_{i=1}^{I}\sum_{t=1}^K \ell(f(\z_{i,t}, \tilde{y}_{i,t}), y_{i,t}) + \lambda r(\bth)
 \end{align}
where $\bth$ contains the parameters of the predictor, $\ell(\cdot)$ is the loss function of a classifier such as hinge loss and cross-entropy loss, $r(\cdot)$ is a regularizer that imposes a certain structure on $\bth$, and $\lambda\geq0$ is a regularization parameter.
In the literature, popular regularizers include $r(\bth) = \|\bth\|_2^2$, $r(\bth) = \|\bth\|_0$, $r(\bth) = \|\bth\|_1$ and $r(\bth)=1_+(\bth)$, i.e., the indicator function of the nonnegative orthant.

For the purpose of illustration, we are particularly interested in but not limited to the SVM classifier. We set $\bth=[\u^T,v]^T$ and $\ell(\cdot)$ as hinge loss, resulting in
\begin{align}\label{hinge}
\min_{\u,v,b} &~ \left\{\frac{1}{IK}\sum_{i=1}^{I}\sum_{t=1}^K\max\(0, 1 - y_{i,t}(\u^T\z_{i,t} + \rho v \tilde{y}_{i,t} + b)\) + \right.\notag\\ 
&\quad\left.\frac{\lambda}{2}\(\|\u\|_2^2+|v|^2\)\right\} \notag\\
\suchthat &~ [\u^T,v]^T\in\mathcal{C}
\end{align}
where $b$ is the intercept, $\mathcal{C}$ denotes a convex set such as $\ell_1$-ball for feature selection and $\rho$ represents the importance of the response at a previous time point on the subsequent one.
In our formulation, the drug response feedback plays an important role and it can be viewed as a ``must-have'' feature. Therefore, the hyper-parameter $\rho$ is critical for the performance of REP. In practice, we recommend to choose a relatively large $\rho$ in an incremental fashion.

\begin{remark}
	The major difference between \eqref{hinge} and the standard SVM-based drug response predictors lies in the feedback $\tilde{y}_{i,t}$: prior art ignored the previous drug response outputs. For the initial time point, there is no response feedback, i.e., $y_{i,0}$ does not exist. So how do we choose $y_{i,0}$? One way could be $y_{i,0} = 0$ which can be interpreted as uncertainty, i.e., we do not know how the patient responds to a treatment. Then we construct the following feedback matrix:
%From Nikos: noted many typos in these two matrices; I corrected; please double-check:
	\begin{align}
	\tilde{\Y}  = \begin{bmatrix}
	\tilde{y}_{1,1} & \tilde{y}_{1,2} & \cdots & \tilde{y}_{1,K} \\
	\tilde{y}_{2,1} & \tilde{y}_{2,2} & \cdots & \tilde{y}_{2,K} \\
	\vdots & & & \\
	\tilde{y}_{I,1} & \tilde{y}_{I,2} & \cdots & \tilde{y}_{I,K}
	\end{bmatrix} %\in \bR^{I\times J}
	\end{align}
	where $\tilde{y}_{i,1} = 0,\forall i=1,\cdots,I$. 
	Finally, the features in the training set is formed by concatenating $\tilde{\Y}$ in $\tZ$ along the gene-axis as shown in the left-bottom corner of Fig. \ref{fig:training}, and the training labels are
	\begin{align}
	\Y  = \begin{bmatrix}
	y_{1,1} & y_{1,2} & \cdots & y_{1,K} \\
	y_{2,1} & y_{2,2} & \cdots & y_{2,K} \\
	\vdots & & & \\
	y_{I,1} & y_{I,2} & \cdots & y_{I,K}
	\end{bmatrix}.
	\end{align}
	
\end{remark}

\subsection{Drug response prediction}
Our method can predict the drug response values for a new patient at any time point. Specifically, given the GELs of a new patient at time $t$, i.e., $\x(t)$, we first check if there are missing values. If so, we employ the factors $\B$ and $\C$ to complete $\x(t)$. Let us denote $\bar\Omega$ and $\bar\Omega^c$ as the sets of indices of the observed and missing elements in $\x(t)$. According to our model in \eqref{xijk}, $\x(t)$ can be uniquely determined by $\B$, $\C$ and an unknown vector $\a$--a latent representation of this new patient. Thus, for the expression level of the $j$th gene at time $t$, we have
\begin{align}\label{xjt}
x_j(t) 
&= 
%\B
\begin{bmatrix}
b_{j1} & \cdots & b_{jF}
\end{bmatrix}
\begin{bmatrix}
a_{1} & & \\ &\ddots & \\ & &a_{F}
\end{bmatrix}
\begin{bmatrix}
c_{t1} \\ \vdots \\ c_{tF}
\end{bmatrix} + \n \notag\\
%\B\diag(\a)(\C(t,:))^T + \n \notag\\
&= \(\C(t,:)\odot\B(j,:)\)\a + n_j,~\forall j\in\bar\Omega
\end{align}
where $n_j$ is the additive noise which is assumed as Gaussian distributed, $\odot$ is the Khatri-Rao (column-wise Kronecker) product, and $\B(j,:)$ and $\C(t,:)$ denote the $t$th row of $\B$ and $\C$, respectively.

Since $\B$ and $\C$ are known, the problem of estimating $\a$ can be formulated as
\begin{align}\label{prob:a}
\hat\a = \arg\min_{\a\geq 0}  \sum_{j\in\bar\Omega} \Big(x_j(t) - \(\C(t,:)\odot\B(j,:)\)\a \Big)^2
\end{align}
which is a nonnegative least squares (NLS) problem and can be optimally solved. 
%\begin{align}
%\hat\a = \(\C(t,:)\odot\B_\Omega\)^\dagger\x_\Omega(t)
%\end{align}
%where $\B_\Omega$ represents the rows of $\B$ that correspond to the indices in $\Omega$, similar to $\x_\Omega(t)$, and $(\cdot)^\dagger$ is the pseudo-inverse. 
We note that to obtain a unique estimate $\hat\a$, the number of available gene expression entries in $\x(t)$ should be $\geq F$.
The GEL vector of the patient is then estimated as
\begin{align}
\g(t) = \(\C(t,:)\odot\B\)\hat\a
\end{align}
which leads to a completed GEL vector
\begin{align}\label{zt}
\z(t) = \cP_{\bar\Omega}(\x(t)) + \cP_{\bar\Omega^c}(\g(t)).
\end{align}
The vector $\z(t)$ together with the cumulated historical drug response $\tilde{y}(t)$, are the input data for our predictor $f(\cdot)$. We estimate the drug response of this patient at time $t$ via
\begin{align}
\hat{y}(t) = f\( \z(t), \tilde{y}(t) \).
\end{align}

\subsubsection{Predicting Unseen GELs}\label{section:futureGEL}
Previously, we have explained how to predict drug response for patients at a certain time point. However, in practice, we are more interested in knowing the drug response of a few time-points in the future from the beginning of a treatment. This requires to know the GELs of all time points up to the time-point of interest {\em a priori}, which is impossible in practice.
In this subsection, we provide an efficient solution that allows to predict the unseen GELs. 

Recall that in our model, the GEL of a patient is determined by three factors, i.e., the latent representation of patient--$\a$, the time evolution factor--$\B$ and the gene factor matrix--$\C$, where $\a$ is different for patients, and needs to be estimated for the new patient. On the other hand, $\B$ and $\C$ are common gene and time evolution bases that reflect different types of patients, as determined from historical patient data -- the training data. Therefore, the problem boils down to the estimation of $\a$ from the initial GELs of the new patient. We can simply substitute $t=1$ in \eqref{prob:a} to find $\hat\a$.
Finally, the GELs for the remaining time points are estimated as
\begin{align}\label{hatxt}
\hat{\x}(t) = \(\C(t,:)\odot\B\)\hat\a, ~\forall t= 2,\cdots,K.
\end{align}

Now we have estimated the unseen GELs for $t\geq 2$, which allows us to predict drug response values for the whole duration of the treatment. We start from $\hat\x(1)$ and estimate the drug response for $t=1$ as
\begin{align}
\hat{y}(1) = f(\hat{\x}(1), \tilde{y}(1))
\end{align}
where $\tilde{y}(1) = 0$. When $\hat{y}(1)$ is available, we set $\tilde{y}(2) = \hat{y}(1)$. With the GEL estimate $\hat{\x}(2)$ from \eqref{hatxt}, we can predict $\hat{y}(2) = f(\hat{\x}(2), \tilde{y}(2))$, and so forth for the other time points.

\begin{remark}
Note that here we substitute {\em predicted} drug responses for the unseen drug responses. Clearly, when actual drug responses for past time ticks are available, they should be used. We only do the substitution here for a preliminary assessment of how well a patient is likely to respond over time, before the beginning of treatment -- which is naturally a more ambitious goal.  
\end{remark}

\section{Simulations}
In this section, we provide some numerical experiments to showcase the effectiveness of REP for drug response prediction from time-course gene expression data. We apply a number of drug response prediction methods including two linear models (EN-LR \cite{fukushima2019elastic} and SVM), one nonlinear model ($K$-nearest neighbor (KNN) \cite{parry2010k}), and a probabilistic graphical model (discriminative loop hidden Markov model (dl-HMM) \cite{lin2008alignment}) to real-world time-course data. We did not include SVM with nonlinear kernels (e.g. Gaussian), since its performance was inferior compared to the linear kernel. Note that EN-LR and dl-HMM were specifically designed for prediction of drug response values based on time-course gene expression data, while SVM and KNN are widely used classification methods.

\subsection{Dataset}
The dataset used is the interferon (IFN)-$\beta$ time-course dataset which is available in the supplementary of \cite{baranzini2004transcription}.
The dataset is collected from 53 Multiple Sclerosis (MS) patients who received IFN-$\beta$ treatment for two years.
The gene expression data (microarray) was obtained from peripheral blood mononuclear cells of the patients, which contained the expression levels of 76 pre-selected genes  over 7 stages (i.e., time-points) of the treatment. However, there are missing values in this dataset, where most missing values were caused by the absence of patients at some stages. Among the patients, only 27 patients have records for all stages, while the other 26 patients missed at least one stage, which resulted in the entire GELs as well as the drug response at that stage being missed. For ease of comparison, we only employ the 27 full records in our experiments, where the final GEL data is with dimension $27\times 7\times 76$ and the response data is with dimension $27\times 7$.

\subsection{Evaluation metric}
We use prediction accuracy (ACC) and area under receiver operating characteristic (ROC) curve (AUC) to evaluate the performance of REP, where ACC is defined as:
\begin{align*}
\mathrm{ACC} = \frac{\rm TP + TN}{\rm TP + FP + FN + TN}
\end{align*}
and ROC curve is created by plotting true positive rate (TPR) versus false positive rate (FPR), defined as
\begin{align*}
\mathrm{TPR} = \frac{\rm TP}{\rm TP + FN} \\
\mathrm{FPR} = \frac{\rm FP}{\rm FP + TN}.
\end{align*}
In the equations above, TP, FP, FN and TN stand for the number of true positives, false positives, false negatives and true negatives, respectively. 
The ACC is calculated using 27-fold cross validation, where in each fold, we select one patient's record as a testing set that contains a $7\times 76$ GEL matrix and a response vector with length 7, while the remaining 26 records are assigned to the training set. 

The hyper-parameters of EN-LR, SVM and KNN are tuned through 5-fold cross validation, while those of REP are tuned slightly differently: we first split the training data into a training set and a validation set consisting of 25 and 1 records, respectively, and then determine the final parameters as those that achieve the highest prediction accuracy in the validation set. It should be noted that the parameters of REP are tuned once but not in a CV way, meaning that REP is tuned not as accurately as the others. For REP, its hyper-parameters include the rank $F$, 
and $\rho$ in \eqref{hinge}, which are selected from $F\in\{2,3,4,5\}$ and $\lambda\in\{0.01,0.1,1,10\}$. 
The SVM method solves the following problem:
\begin{align}
\min_{\u,b} &~ \frac{1}{IK}\sum_{i=1}^{I}\sum_{t=1}^K\max\(0, 1 - y_{i,t}(\u^T\z_{i,t} + b)\) + \frac{\lambda}{2}\|\u\|_2^2
\end{align}
where its parameter $\lambda$ is tuned from $\{0.01,0.1,1,10\}$; for EN-LR, we set $\alpha=0.5$ which is a hyper-parameter balancing the ridge and LASSO regularizations; for KNN, the number of neighbors is selected from $\{3,5,8,10\}$.
After that, we apply the trained classifier to the testing data to calculate ACC. We implemented REP, EN-LR, SVM and KNN in Python 3.7. Since the authors of dl-HMM have published their MATLAB codes (\url{http://www.cs.cmu.edu/~thlin/tram/}), we implemented these two methods in MATLAB. For both HMM methods, we choose the number of hidden states as 4 which is as  suggested in \cite{lin2008alignment}.

\subsection{Results}

\begin{figure}[t]
	\centering
	\includegraphics[trim=2.5cm 5.0cm 2.5cm 3.5cm, width=8cm]{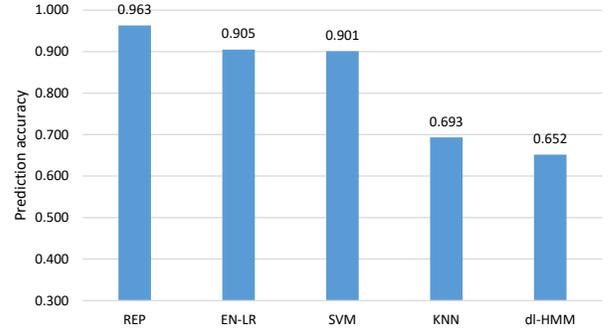}
	\caption{Prediction accuracy comparison on raw data, where the percentage of missing values is 0.23\%.}
	\label{fig:accraw}
\end{figure}

%\begin{figure*}
%	\centering
%	\includegraphics[trim=3cm 7cm 3cm 6cm, width=12cm]{plot/acc_vs_perc_missing}
%	\caption{Performance comparison on prediction accuracy as a function of percentage of missing values.}
%	\label{fig:accvspercmissing}
%\end{figure*}

% Table generated by Excel2LaTeX from sheet 'Sheet1'

\begin{table}[htbp]
  \centering
  \caption{ACC \& AUC versus percentage of missing values}
    \begin{tabular}{c|c|c|c|c|c|c}
    \toprule
    \multicolumn{1}{c|}{Metrics} & \multicolumn{1}{c|}{$\%$ miss} & \multicolumn{1}{c|}{REP} & \multicolumn{1}{c|}{EN-LR} & \multicolumn{1}{c|}{SVM} & \multicolumn{1}{c|}{KNN} & \multicolumn{1}{c}{dl-HMM} \\
    \midrule
    \multicolumn{1}{c|}{\multirow{4}[8]{*}{ACC}} & 5  & \textbf{94.8} & 90.5 & 88.4 & 70.4 & 65.1 \\
    \cmidrule{2-7}          & 10   & \textbf{92.6} & 88.4 & 88.7 & 69.3 & 59.6 \\
    \cmidrule{2-7}          & 15  & \textbf{89.4} & 85.2 & 81.6 & 64.3 & 59.6 \\
    \cmidrule{2-7}          & 20   & \textbf{88.5} & 84.7 & 81.5 & 66.8 & 55.6 \\
    \midrule
    \midrule
    \multicolumn{1}{c|}{\multirow{4}[8]{*}{AUC}} & 5  & \textbf{96.9} & 93.0  & 91.2 & 50.2 & 53.9 \\
    \cmidrule{2-7}          & 10   & \textbf{96.8} & 90.1 & 88.6 & 49.4 & 54.7 \\
    \cmidrule{2-7}          & 15  & \textbf{95.4} & 87.4 & 84.3 & 51.9 & 54.7 \\
    \cmidrule{2-7}          & 20   & \textbf{93.2} & 88.1 & 87.1 & 52.3 & 56.4 \\
    \bottomrule
    \end{tabular}%
  \label{tab:addlabel}%
\end{table}%

\begin{figure}
	\centering
	\includegraphics[width=8cm]{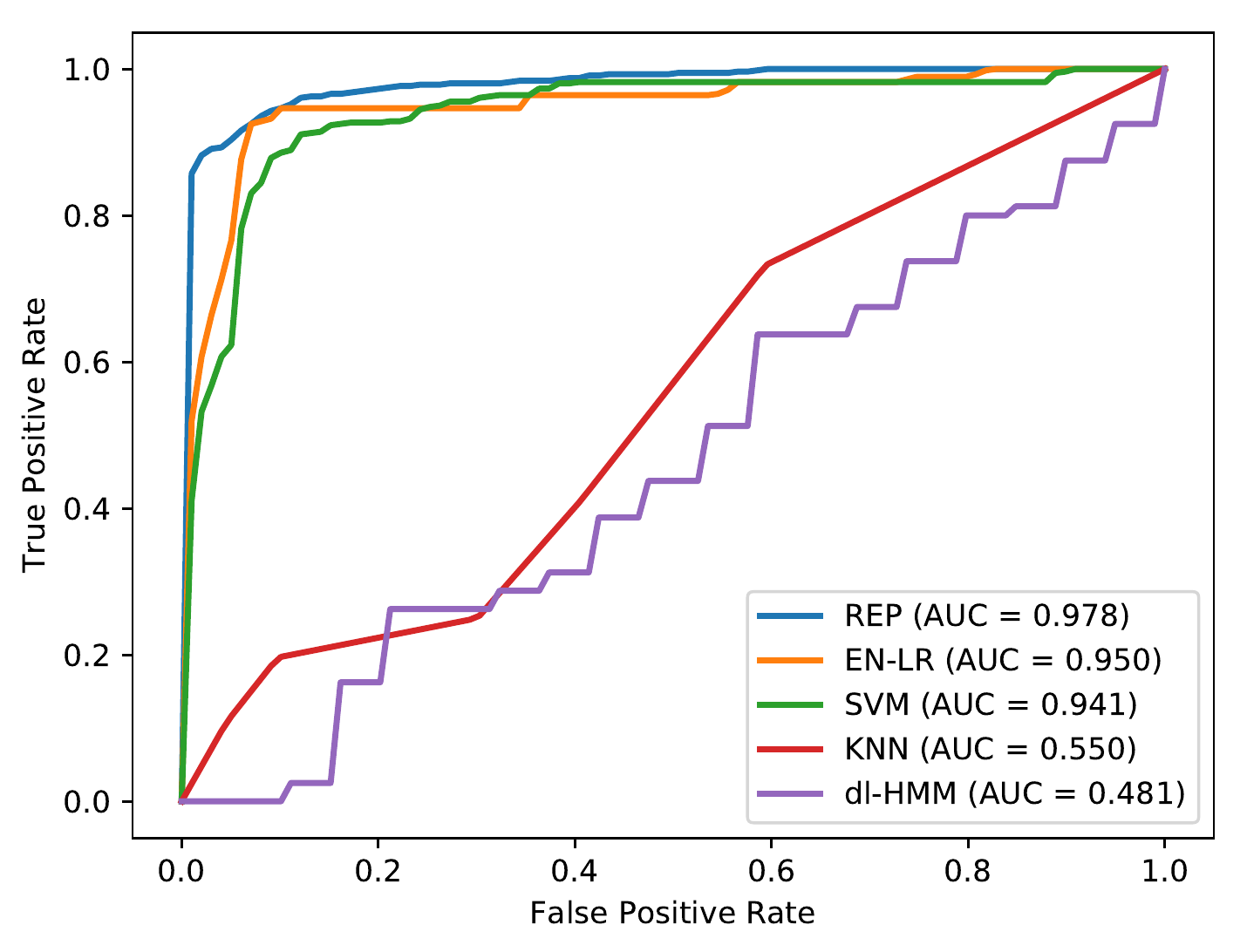}
	\caption{ROC curves.}
	\label{fig:rocraw1}
\end{figure}

Fig. \ref{fig:accraw} shows the prediction accuracy of the five algorithms. In this example, the missing values is only 0.23\%. For all the methods, the missing GELs were completed using nonnegative tensor completion in Section II-A. It can be seen that REP achieves a higher prediction accuracy compared to other methods and its performance is followed by the EN-LR and SVM algorithms. The KNN and dl-HMM algorithms do not perform well. We observed that dl-HMM was very sensitive to the data, and it frequently produced not-a-number (NaN) values. Particularly for patients with ID 1117161, 1215133 and 995640, the original codes of dl-HMM always produced NaN of the three patients' responses. This is the main reason why dl-HMM had low prediction accuracy. 
We note that REP takes a similar formulation as the SVM. However, REP is 6.2\% more accurate than SVM, which implies that the recursive structure in REP is helpful in improving the prediction accuracy. In Fig. \ref{fig:rocraw1}, the ROC curves are plotted, which shows that the proposed method has the highest AUC value.

Next, we sought to determine the effect of missing values on the performance of these methods. For this purpose, we randomly sampled the GEL data and hid the selected entries. As the percentage of missing values increases, we can see in Table  \ref{tab:addlabel} that all methods suffer performance loss, but REP's ACC and AUC remain the highest in all cases. We highlight that when the percentage of missing values is smaller than 15\%, REP has ACC close to 90\% and AUC greater than 95\%. EN-LR outperforms the classical SVM method in many cases. When the percentage of missing values increases, the performance of EN-LR and SVM drops significantly, while that of REP still remains at a high level. For example, when percentage of missing increases to 15\%, the AUC of EN-LR drops to 87.4\%, but ours is 95.4\%, which indicates that REP is more robust against missing values.

\begin{figure}
	\centering
	\includegraphics[trim=2cm 5cm 1cm 4cm, width=9cm]{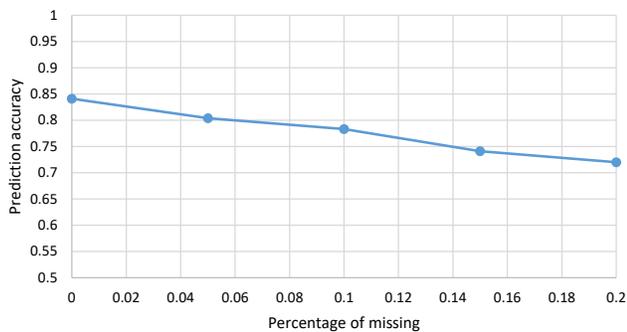}
	\caption{Prediction accuracy of REP with estimated GEL.}
	\label{fig:resultpredgelestreponly}
\end{figure}

The above examples were conducted under the full knowledge of GELs for each time point.  However, we are more interested in predicting the drug response in the beginning of a treatment, meaning that we only have a few GELs  at the first time point while all the subsequent GELs are unavailable. Specifically, we can first learn factors $\{\B,\C\}$ using nonnegative tensor completion, then follow the procedures in Section \ref{section:futureGEL} to predict the GELs, and finally, use the estimated GELs for drug response prediction. To show the effectiveness of REP, we only used the GELs of new patients at the first time point, i.e., $t=1$, while hiding the GELs for other time points through $t=2$ to $7$.  We studied the performance of REP where only GELs for the first time-points 
%From Nikos: very confusing: from a few or only the first time-point?
were available. Fig. \ref{fig:resultpredgelestreponly} shows the prediction accuracy of REP by using estimated GELs, where the percentage of missing stands for the missing values in the training set. We see that REP with estimated GELs achieves about 85\% accuracy. Although this performance is not as good as REP with the ground truth GELs (see Fig. 2), we must emphasize that such performance is obtained by only knowing the GELs from the first stage of the treatment, which is more meaningful in practice.

\section{Conclusion}
We studied the problem of drug response prediction for time-course gene expression data and presented a computational algorithm (REP) that: (i) has a recursive structure that integrates past drug response records for subsequent predictions, (ii) offers higher prediction accuracy than several classical algorithms such as SVM and LR, (iii) exploits the tensor structure of the data for missing GEL completion and unseen GEL prediction, (iv) can predict drug response of different stages of a treatment from some initial GEL measurements. The achieved performance improvement for real data application suggests that our method serves as a better predictor of drug response using the time-course data.

\subsection*{}
\noindent \textit{Conflict of Interest}: none declared

%\bibliographystyle{natbib}  
%\bibliographystyle{achemnat}
%\bibliographystyle{plainnat}
%\bibliographystyle{abbrv}
%\bibliographystyle{bioinformatics}
%\bibliographystyle{plain}
%\bibliography{mybib}

\bibliographystyle{IEEEtran}
\bibliography{IEEEabrv,mybib}

\end{document}